\documentclass[10pt,twocolumn,letterpaper]{article}

\usepackage[utf8]{inputenc}
\usepackage{cvpr}
\usepackage{times}
\usepackage{epsfig}
\usepackage{graphicx}
\usepackage{amsmath}
\usepackage{amssymb}
\usepackage{amsfonts,amsthm}
\usepackage{color}
\usepackage{caption}
\usepackage{subcaption}
\usepackage{xspace}
\usepackage{array}
\usepackage{cite}
\usepackage{placeins}
\usepackage{mathtools}
\usepackage{booktabs}
\usepackage[pagebackref=true,breaklinks=true,letterpaper=true,colorlinks,bookmarks=false]{hyperref}

\cvprfinalcopy %
\def\cvprPaperID{5290} %

\ifcvprfinal\fi

\newcommand{\bD}{\mathbf{D}}

\newcommand{\bg}{\mathbf{g}}

\newcommand{\bI}{\mathbf{I}}

\newcommand{\bM}{\mathbf{M}}
\newcommand{\bn}{\mathbf{n}}

\newcommand{\bp}{\mathbf{p}}
\newcommand{\bq}{\mathbf{q}}
\newcommand{\br}{\mathbf{r}}

\newcommand{\bt}{\mathbf{t}}
\newcommand{\bu}{\mathbf{u}}

\newcommand{\bw}{\mathbf{w}}
\newcommand{\bx}{\mathbf{x}}

\newcommand{\bz}{\mathbf{z}}

\newcommand{\nR}{\mathbb{R}}

\newcommand{\cL}{\mathcal{L}}

\newcommand{\cP}{\mathcal{P}}

\newcommand{\cZ}{\mathcal{Z}}

\newcommand{\figref}[1]{Fig.~\ref{#1}}
\newcommand{\secref}[1]{Section~\ref{#1}}

\newcommand{\tabref}[1]{Table~\ref{#1}}

\DeclareMathOperator*{\argmin}{argmin~}

\makeatletter
\DeclareRobustCommand\onedot{\futurelet\@let@token\@onedot}
\def\@onedot{\ifx\@let@token.\else.\null\fi\xspace}
\def\eg{e.g\onedot} 
\def\ie{i.e\onedot}

\def\wrt{wrt\onedot}

\def\etal{et~al\onedot}

\makeatother

\newcommand{\boldparagraph}[1]{\vspace{0.2cm}\noindent{\bf #1:}}

\definecolor{darkgreen}{rgb}{0,0.7,0}

\newcommand{\texnet}{\mathbf{t}}
\newcommand{\onet}{f}
\newcommand{\imggt}{\bI}

\newcommand{\imgpred}{\mathbf{\hat{I}}}
\newcommand{\depthpred}{\hat{d}}
\newcommand{\pointpred}{\mathbf{\hat{p}}}

\newcolumntype{L}[1]{>{\raggedright\let\newline\\\arraybackslash\hspace{0pt}}m{#1}}
\newcolumntype{C}[1]{>{\centering\let\newline\\\arraybackslash\hspace{0pt}}m{#1}}
\newcolumntype{R}[1]{>{\raggedleft\let\newline\\\arraybackslash\hspace{0pt}}m{#1}}

\renewcommand*{\thefootnote}{\fnsymbol{footnote}}

\begin{document}
 
\title{Differentiable Volumetric Rendering:\\Learning Implicit 3D Representations without 3D Supervision}
\author{Michael Niemeyer$^{1,2}$ \quad Lars Mescheder$^{1,2,3}\footnotemark[2]$ \quad Michael Oechsle$^{1,2,4}$ \quad Andreas Geiger$^{1,2}$\\
$^1$Max Planck Institute for Intelligent Systems, Tübingen \qquad $^2$University of Tübingen\\
$^3$Amazon, Tübingen \qquad $^4$ETAS GmbH, Bosch Group, Stuttgart\\
{\tt\small \{firstname.lastname\}@tue.mpg.de}
}

\maketitle

\footnotetext[2]{This work was done prior to joining Amazon.}
\renewcommand*{\thefootnote}{\arabic{footnote}}
 \setcounter{footnote}{0}

\begin{abstract}
Learning-based 3D reconstruction methods have shown impressive results. However, most methods require 3D supervision which is often hard to obtain for real-world datasets. Recently, several works have proposed differentiable rendering techniques to train reconstruction models from RGB images. Unfortunately, these approaches are currently restricted to voxel- and mesh-based representations, suffering from discretization or low resolution. In this work, we propose a differentiable rendering formulation for implicit shape and texture representations. Implicit representations have recently gained popularity as they represent shape and texture continuously. Our key insight is that depth gradients can be derived analytically using the concept of implicit differentiation. This allows us to learn implicit shape and texture representations directly from RGB images. We experimentally show that our single-view reconstructions rival those learned with full 3D supervision. Moreover, we find that our method can be used for multi-view 3D reconstruction, directly resulting in watertight meshes.
\end{abstract}

\begin{figure}
	\centering
	\resizebox{1.\linewidth}{!}{
	\includegraphics[width=1.\textwidth]{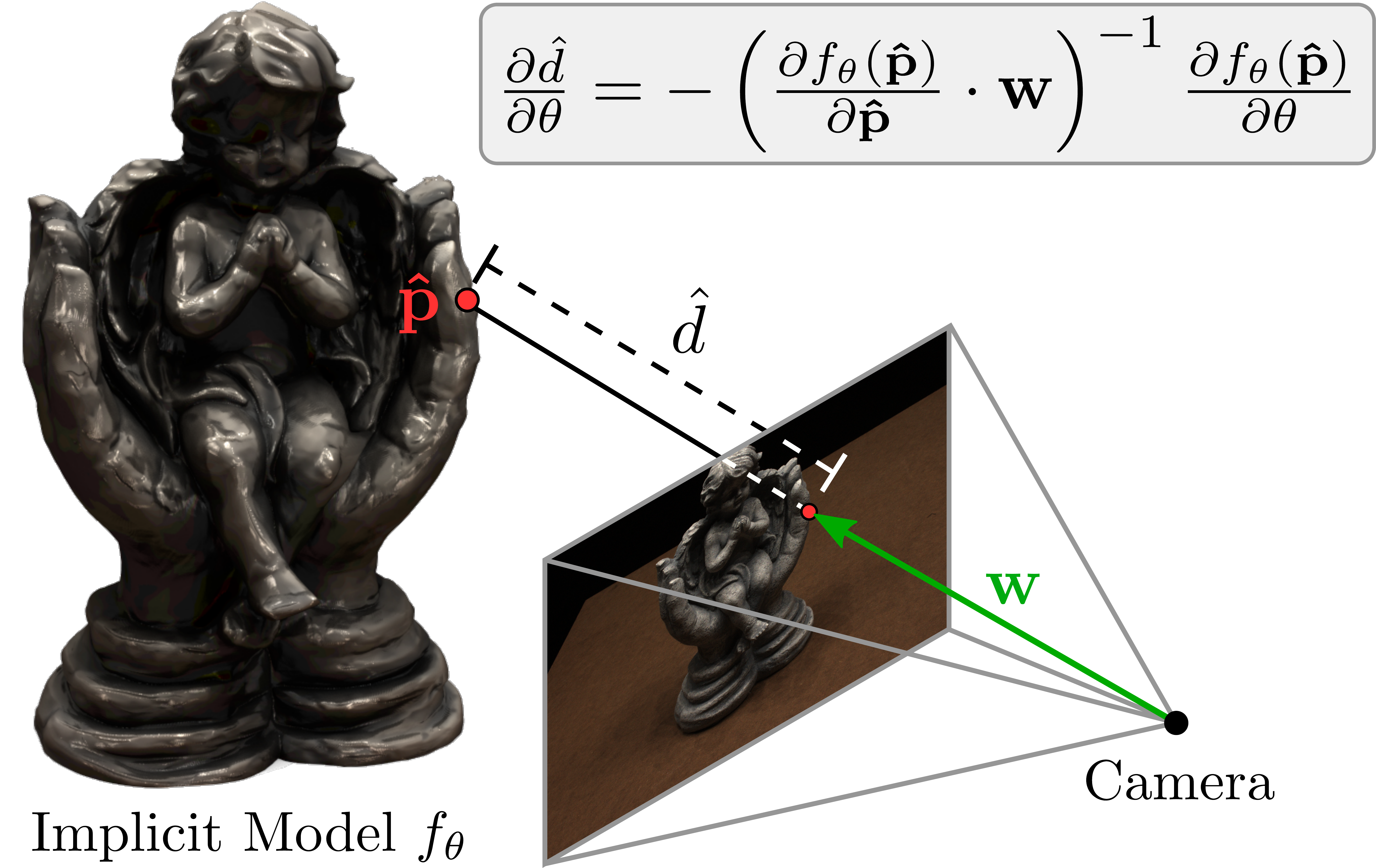}
	}
	\caption{
		\textbf{Overview.}
		We show that volumetric rendering is inherently differentiable for implicit shape and texture representations.
		Using an analytic expression for the gradient of the depth $\frac{\partial \hat{d}}{\partial \theta}$ \wrt the network parameters $\theta$, we learn implicit 3D representations $f_\theta$ from 2D images.
		}
	\label{fig:teaser}
	\vspace{-0.5cm}
\end{figure}

\section{Introduction}

In recent years, learning-based 3D reconstruction approaches have achieved impressive results \cite{Mescheder2019CVPR, Park2019CVPR, Chen2018CVPR, Choy2016ECCV, Fan2017CVPR, Groueix2018CVPR, Liao2018CVPR, Riegler2017CVPR, Wang2018ECCVc, Michalkiewicz2019ICCV}.
By using rich prior knowledge obtained during the training process, they are able to infer a 3D model from as little as a single image.
However, most learning-based methods are restricted to synthetic data, mainly because they require accurate 3D ground truth models as supervision for training.

To overcome this barrier, recent works have investigated approaches that require only 2D supervision in the form of depth maps or multi-view images. 
Most existing approaches achieve this by modifying the rendering process to make it differentiable \cite{Baumgart1974SU,Paschalidou2018CVPR, Tulsiani2017CVPR, LIU2019ICCV, Zienkiewicz2016IROS, Liu2017ICCVabc, Loper2014ECCV, KATO2018CVPR, Genova2018CVPR, Phuoc2018NIPS, Kundu2018CVPR, Deschaintre2018TOC, Tewari2017ICCV, Tewari2018CVPR, Richardson2017CVPR, Petersen2019ARXIV, Chen2019NIPS}.
While yielding compelling results, they are restricted to specific 3D representations (\eg voxels or meshes) that suffer from discretization artifacts and the computational cost limits them to small resolutions or deforming a fixed template mesh.
At the same time, implicit representations \cite{Mescheder2019CVPR, Park2019CVPR, Chen2018CVPR} for shape and texture \cite{Oechsle2019ICCV,Saito2019ICCV} have been proposed which do not require discretization during training and have a constant memory footprint.
However, existing approaches using implicit representations require 3D ground truth for training and it remains unclear how to learn implicit representations from image data alone.

\vspace{0.1cm}\noindent\textbf{Contribution:} %
In this work, we introduce \emph{Differentiable Volumetric Rendering (DVR)}.
Our key insight is that we can derive analytic gradients for the predicted depth map with respect to the network parameters of the implicit shape and texture representation (see~\figref{fig:teaser}).
This insight enables us to design a differentiable renderer for implicit shape and texture representations and allows us to learn these representations solely from multi-view images and object masks.
Since our method does not have to store volumetric data in the forward pass, its memory footprint is independent of the sampling accuracy of the depth prediction step.
We show that our formulation can be used for various tasks such as single- and multi-view reconstruction, and works with synthetic and real data.
In contrast to \cite{Oechsle2019ICCV}, we do not need to condition the texture representation on the geometry, but learn a \textit{single model} with shared parameters that represents both geometry and texture.
Our code and data are provided at \url{https://github.com/autonomousvision/differentiable_volumetric_rendering}.

\section{Related Work}

\boldparagraph{3D Representations}
Learning-based 3D reconstruction approaches can be categorized \wrt the representation they use as voxel-based~\cite{Choy2016ECCV,Brock2016ARXIV, Wu2016NIPS, Rezende2016NIPS,Gadelha2017THREEDV,Stutz2018CVPR, Riegler2017CVPR,Xie2019ICCV}, point-based~\cite{Fan2017CVPR, Achlioptas2018ICML, Thomas2019ICCV, Yang2019ICCV, Jiang2018ECCV, Li2018ECCVc}, 
mesh-based\cite{Liao2018CVPR, Groueix2018CVPR, Wang2018ECCVc, Kanazawa2018ECCV, Pan2019ICCV}, or implicit representations~\cite{Mescheder2019CVPR, Chen2018CVPR, Park2019CVPR,Wang2019NIPS, Saito2019ICCV,  Michalkiewicz2019ICCV, Genova2019ICCV, HuangLi2018ECCV, Atzmon2019NIPS}.

Voxels can be easily processed by standard deep learning architectures, but even when operating on sparse data structures \cite{Graham2015BMVC, Riegler2017CVPR, Tatarchenko2017ICCV}, they are limited to relatively small resolution.
While point-based approaches \cite{Fan2017CVPR, Achlioptas2018ICML, Thomas2019ICCV, Yang2019ICCV, Li2018ECCVc} are more memory-efficient, they require intensive post-processing because of missing connectivity information.
Most mesh-based methods do not perform post-processing, but they often require a deformable template mesh \cite{Wang2018ECCVc} or represent geometry as a collection of 3D patches \cite{Groueix2018CVPR} which leads to self-intersections and non-watertight meshes.

To mitigate these problems, implicit representations have gained popularity \cite{Mescheder2019CVPR, Chen2018CVPR, Park2019CVPR,Wang2019NIPS, Saito2019ICCV,  Michalkiewicz2019ICCV, Genova2019ICCV, HuangLi2018ECCV, Niemeyer2019ICCV, Oechsle2019ICCV, Atzmon2019NIPS}.
By describing 3D geometry and texture implicitly, \eg, as the decision boundary of a binary classifier \cite{Mescheder2019CVPR,Chen2018CVPR}, they do not discretize space and have a fixed memory footprint.

In this work, we show that the volumetric rendering step for implicit representations is inherently differentiable.
In contrast to previous works, this allows us to learn implicit 3D shape and texture representations using 2D supervision. 

\boldparagraph{3D Reconstruction}
Recovering 3D information which is lost during the image capturing process is one of the long-standing goals of computer vision \cite{Hartley2003CUP}.
Classic multi-view stereo (MVS) methods \cite{Seitz2006CVPR,Bleyer2011BMVC,Galliani2016,Schoenberger2016ECCV,Bonet1999ICCV,Broadhurst2001ICCV,Kutulakos2000IJCV,Seitz1997CVPR,Prock1998} usually match features between neighboring views~\cite{Bleyer2011BMVC,Galliani2016,Schoenberger2016ECCV} or reconstruct the 3D shape in a voxel grid \cite{Bonet1999ICCV,Broadhurst2001ICCV,Kutulakos2000IJCV, Seitz1997CVPR,Prock1998}.
While the former methods produce depth maps as output which have to be fused in a lossy post-processing step, \eg, using volumetric fusion \cite{Curless1996SIGGRAPH}, the latter approaches are limited by the excessive memory requirements of 3D voxel grids.
In contrast to these highly engineered approaches, our generic method directly outputs a consistent representation in 3D space which can be easily converted into a watertight mesh while having a constant memory footprint.

Recently, learning-based approaches \cite{Riegler2017THREEDV,Paschalidou2018CVPR, Donne2019CVPR, Leroy2018ECCV, Pohan2018CVPR, Yao2018ECCV, Yao2019CVPR} have been proposed that either learn to match image features \cite{Leroy2018ECCV}, refine or fuse depth maps \cite{Donne2019CVPR,Riegler2017THREEDV}, optimize parts of the classical MVS pipeline \cite{Paschalidou2019CVPR}, or replace the entire MVS pipeline with neural networks that are trained end-to-end  \cite{Pohan2018CVPR,Yao2018ECCV, Yao2019CVPR}.
In contrast to these learning-based approaches, our method can be supervised from 2D images alone and outputs a consistent 3D representation.

\boldparagraph{Differentiable Rendering}
We focus on methods that learn 3D geometry via differentiable rendering in contrast to recent neural rendering approaches~\cite{Sitzmann2019CVPR,Phuoc2019ICCV,Phuoc2020ARXIV,Liao2020CVPR} which synthesize high-quality novel views but do not infer the 3D object.
They can again be categorized by the underlying representation of 3D geometry that they use.

Loper~\etal~\cite{Loper2014ECCV} propose OpenDR which approximates the backward pass of the traditional mesh-based graphics pipeline and has inspired several follow-up works \cite{KATO2018CVPR,LIU2019ICCV, Zienkiewicz2016IROS, Genova2018CVPR, Chen2019NIPS, Henderson2018BMVC, Henderson2019IJCV}.
Liu \etal~\cite{LIU2019ICCV} replace the rasterization step with a soft version to make it differentiable. 
While yielding compelling results in reconstruction tasks, these approaches require a deformable template mesh for training, restricting the topology of the output.

Another line of work operates on voxel grids \cite{Paschalidou2019CVPR, Tulsiani2017CVPR, Lombardi2019,Phuoc2018NIPS}.
Paschalidou \etal~\cite{Paschalidou2019CVPR} and Tulsiani \etal~\cite{Tulsiani2017CVPR} propose a probabilistic ray potential formulation. 
While providing a solid mathematical framework, all intermediate evaluations need to be saved for backpropagation, restricting these approaches to relatively small-resolution voxel grids.

Liu~\etal~\cite{Liu2019NIPS} propose to infer implicit representations from multi-view silhouettes by performing max-pooling over the intersections of rays with a sparse number of supporting regions. %
In contrast, we use texture information enabling us to improve over the visual hull and to reconstruct concave shapes.
Sitzmann~\etal~\cite{Sitzmann2019NIPS} infer implicit scene representations from RGB images via an LSTM-based differentiable renderer. 
While producing high-quality renderings, the geometry cannot be extracted directly and intermediate results need to be stored for computing gradients.
In contrast, we show that volumetric rendering is \textit{inherently differentiable} for implicit representations. Thus, no intermediate results need to be saved for the backward pass.

\section{Method}
\begin{figure*}
    \centering
    \includegraphics[width=\linewidth]{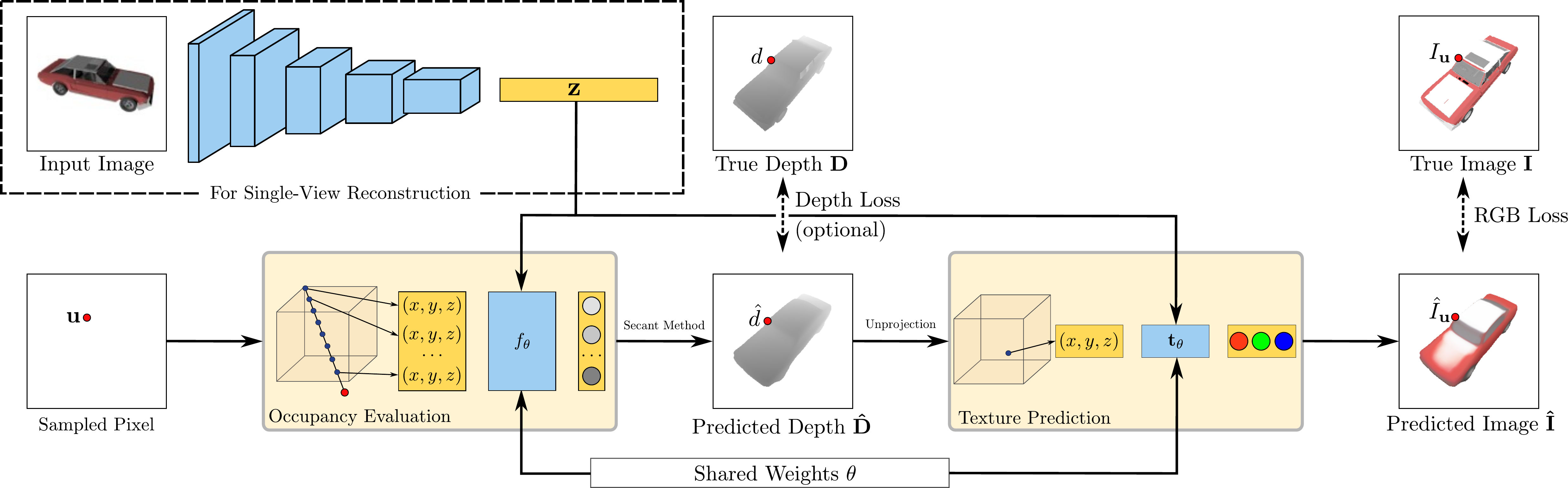}
    \caption{
        \textbf{Differentiable Volumetric Rendering.} 
        We first predict the surface depth $\hat d$ by performing occupancy evaluations for a given camera matrix.
        To this end, we project sampled pixel~$\bu$ to 3D and evaluate the occupancy network at fixed steps on the ray cast from the camera origin towards this point.
        We then unproject the surface depth into 3D and evaluate the texture field at the given 3D location.
        The resulting 2D rendering $\imgpred$ can be compared to the ground truth image.
        When we also have access to ground truth depth maps, we can define a loss directly on the predicted surface depth.
        We can make our model conditional by incorporating an additional image encoder that predicts a global descriptor $\bz$ of both shape and texture.
        \vspace{-0.35cm}
        }
    \label{fig:overview}
\end{figure*}

In this section, we describe our Differentiable Volumetric Rendering (DVR) approach. 
We first define the implicit neural representation which we use for representing 3D shape and texture.
Next, we provide a formal description of DVR and all relevant implementation details.
An overview of our approach is provided in \figref{fig:overview}.

\subsection{Shape and Texture Representation}

\vspace{-0.25cm}
\boldparagraph{Shape}
In contrast to discrete voxel- and point-based representations, we represent the 3D shape of an object \emph{implicitly} using the occupancy network introduced in \cite{Mescheder2019CVPR}: %
\begin{equation}
    \onet_\theta: \mathbb{R}^3 \times \cZ \to  [0, 1]
\end{equation}
An occupancy network $\onet_\theta(\bp,\bz)$ assigns a probability of occupancy
to every point $\bp \in \nR^3$ in 3D space.
For the task of single-view reconstruction, we process the input image with an encoder network $\bg_\theta(\cdot)$ and use the output $\bz \in \cZ$ to condition $\onet_\theta$.
The 3D surface of an object is implicitly determined by the level set $f_\theta = \tau$ for a threshold parameter $\tau \in [0, 1]$ and can be extracted at arbitrary resolution using isosurface extraction techniques.\footnote{%
See Mescheder \etal \cite{Mescheder2019CVPR} for details.
}

\boldparagraph{Texture}
Similarly, we can describe the texture of a 3D object using a texture field \cite{Oechsle2019ICCV}
\begin{equation}
    \texnet_\theta: \nR^3 \times \cZ \to  \nR^3
\end{equation}
which regresses an RGB color value for every point ${\bp \in \nR^3}$ in 3D space.
Again, $\texnet_\theta$ can be conditioned on a latent embedding $\bz$ of the object.
The texture of an object is given by the values of $\texnet_\theta$ on the object's surface ($f_\theta = \tau$).
In this work, we implement $\onet_\theta$ and $\texnet_\theta$  
as a single neural network with two shallow heads.

\boldparagraph{Supervision}
Recent works \cite{Mescheder2019CVPR,Chen2018CVPR,Park2019CVPR,Oechsle2019ICCV,Saito2019ICCV} have shown that it is possible to learn $\onet_\theta$ and $\texnet_\theta$ with 3D supervision (\ie, ground truth 3D models).
However, ground truth 3D data is often very expensive or even impossible to obtain for real-world datasets.
In the next section, we introduce DVR, an alternative approach that enables us to learn both $\onet_\theta$ and $\texnet_\theta$ from 2D images alone.
For clarity, we drop the condition variable $\bz$ in the following.

\subsection{Differentiable Volumetric Rendering}
\begin{figure}
    \centering
    \includegraphics[width=.8\linewidth]{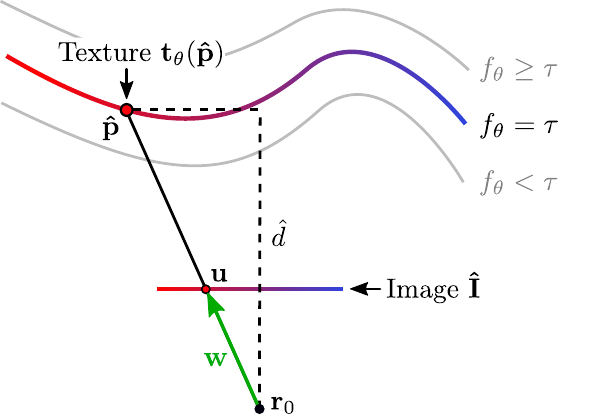}
    \caption{
        \textbf{Notation.} 
        To render an object from the occupancy network $\onet_\theta$ and texture field $\texnet_\theta$, 
        we cast a ray with direction $\bw$ through a pixel $\bu$ and determine the intersection point $\pointpred$ with 
        the isosurface $\onet_\theta(\bp) = \tau$.
        Afterwards, we evaluate the texture field $\texnet_\theta$ at $\pointpred$ to obtain the color prediction $\imgpred_\bu$ at $\bu$.
        }
    \label{fig:notation}
\end{figure}

Our goal is to learn $\onet_\theta$ and $\texnet_\theta$ from 2D image observations.
Consider a single image observation. We define a photometric reconstruction loss
\begin{equation}\label{eq:loss}
\cL(\imgpred, \imggt) = \sum_\bu {\|\imgpred_\bu - \imggt_\bu \|}
\end{equation}
which we aim to optimize.
Here, $\imggt$ denotes the observed image and $\imgpred$ is the image rendered by our implicit model.\footnote{Note that the rendered image $\imgpred$ depends on $\theta$ through $f_\theta$ and $\bt_\theta$.
We have dropped this dependency here to avoid clutter in the notation.}
Moreover, $\bI_\bu$ denotes the RGB value of the observation $\imggt$ at pixel $\bu$ and $ {\|\cdot\|}$ is a (robust) photo-consistency measure such as the $\ell_1$-norm.
To minimize the reconstruction loss $\cL$ \wrt the network parameters $\theta$ using gradient-based optimization techniques, we must be able to (i) \textbf{render} $\imgpred$ given $\onet_\theta$ and $\texnet_\theta$ and (ii) compute \textbf{gradients} of $\cL$ \wrt the network parameters $\theta$.
Our core contribution is to provide solutions to both problems, leading to an efficient algorithm for learning implicit 3D representations from 2D images.

\boldparagraph{Rendering}
For a camera located at $\br_0$ we
can predict the color $\imgpred_\bu$ at pixel $\bu$ by casting a ray from $\br_0$ through $\bu$ and determining the first point of intersection $\pointpred$ with the isosurface
$\{\bp \in \mathbb{R}^3 | \onet_\theta(\bp) = \tau\}$
as illustrated in \figref{fig:notation}.
The color value $\imgpred_\bu$ is then given by $\imgpred_\bu = \texnet_\theta(\pointpred)$.
We refer the reader to \secref{sec:implementation} for details on the ray casting process.

\boldparagraph{Gradients}
To obtain gradients of $\cL$ with respect to $\theta$, we first use the multivariate chain rule:
\begin{equation}\label{eq:dL_dtheta}
    \frac{\partial\cL}{\partial \theta}
    = 
    \sum_\bu
    \frac{\partial\cL}{\partial \imgpred_\bu}
    \cdot
    \frac{\partial \imgpred_\bu}{\partial \theta}
\end{equation}
Here, $\frac{\partial  \bg}{\partial \bx}$ denotes the Jacobian matrix for a vector-valued function $\bg$ with vector-valued argument $\bx$ and $\cdot$ indicates matrix multiplication.
By exploiting $\imgpred_\bu = \texnet_\theta(\pointpred)$, we obtain
\begin{equation}\label{eq:dIu_dtheta}
    \frac{\partial \imgpred_\bu}{\partial \theta}
    = 
    \frac{\partial \texnet_\theta(\hat{\bp})}{\partial \theta}
    + \frac{\partial \texnet_\theta(\hat{\bp})}{\partial \hat{\bp}}
    \cdot
    \frac{\partial \hat{\bp}}{\partial \theta}
\end{equation}
since both $\bt_\theta$ as well as $\hat{\bp}$ depend on $\theta$.
Because $\pointpred$ is defined implicitly, calculating $\tfrac{\partial \hat{\bp}}{\partial \theta}$ is non-trivial.
We first exploit that $\pointpred$ lies on the ray from $\br_0$ through $\bu$. For any pixel $\bu$, this ray can be described by $\br(d) = \br_0 + d \bw$ where $\bw$ is the vector connecting $\br_0$ and $\bu$ (see \figref{fig:notation}).
Since $\pointpred$ must lie on $\br$, there exists a depth value $\hat d$, such that $\pointpred = \br(\hat d)$.
We call $\hat d$ the \emph{surface depth}.
This enables us to rewrite $\tfrac{\partial \hat{\bp}}{\partial \theta}$ as
\begin{equation}\label{eq:dphat_dtheta}
    \frac{\partial \hat{\bp}}{\partial \theta}
    =
    \frac{\partial\br(\hat d)}{\partial \theta}
    =
    \bw \frac{\partial \hat d}{\partial \theta} 
\end{equation}
For computing the gradient of the surface depth $\hat d$ with respect to $\theta$ we exploit \textit{implicit differentiation} \cite{Atzmon2019NIPS, Rudin1964}.
Differentiating $\onet_\theta( \hat{\bp}) = \tau$ on both sides \wrt $\theta$, we obtain:
\begin{equation}\label{eq:implicit_diff}
\begin{split}
\frac{\partial \onet_\theta(\hat{\bp})}{\partial \theta} + \frac{\partial \onet_\theta(\hat{\bp})}{\partial \hat{\bp}} \cdot \frac{\partial \hat{\bp}}{\partial \theta}
&= 0 
\\
\xLeftrightarrow{} %
\frac{\partial \onet_\theta(\hat{\bp})}{\partial \theta} +
\frac{\partial \onet_\theta(\hat{\bp})}{\partial \hat{\bp}} 
\cdot 
\bw
\frac{\partial \hat d}{\partial \theta} 
&= 0%
\end{split}
\end{equation}
Rearranging \eqref{eq:implicit_diff}, we arrive at the following closed form expression for the gradient of the surface depth $\hat{d}$:
\begin{equation}\label{eq:dhatd_dtheta}
\frac{\partial \hat d}{\partial \theta}
=
-\left(\frac{\partial \onet_\theta(\hat{\bp})}{\partial \hat{\bp}} \cdot \bw\right)^{-1} 
\frac{\partial \onet_\theta(\hat{\bp})}{\partial \theta}
\end{equation}
We remark that calculating the gradient of the surface depth $\hat d$ \wrt the network parameters $\theta$ only involves calculating the gradient of $\onet_\theta$ at $\hat{\bp}$ \wrt the network parameters $\theta$ and the surface point $\hat{\bp}$. 
Thus, in contrast to voxel-based approaches \cite{Tulsiani2017CVPR,Paschalidou2018CVPR}, we do not have to store intermediate results %
(\eg, volumetric data)
for computing the gradient of the loss \wrt the parameters, resulting in a memory-efficient algorithm.
In the next section, we describe our implementation of DVR which makes use of reverse-mode automatic differentiation to compute the full gradient \eqref{eq:dL_dtheta}.

\subsection{Implementation}\label{sec:implementation}
To use automatic differentiation, we have to implement the forward and backward pass for the surface depth prediction step $\theta \to \hat d$.
In the following, we describe how both passes are implemented.
For more details, we refer the reader to the supplementary material.

\boldparagraph{Forward Pass}
\label{subsec:forward_pass}
As visualized in \figref{fig:notation}, we can determine $\hat d$ by finding the first occupancy change on the ray $\br$.
To detect an occupancy change, we evaluate the occupancy network $\onet_\theta(\cdot)$ at $n$ equally-spaced samples on the ray $\{\bp_j^\text{ray}\}_{j=1}^n$.
Using a step size of $\Delta s$, we can express the coordinates of these point in world-coordinates as
\begin{equation}
	\bp_j^\text{ray} = \br(j \Delta s+s_0) %
\end{equation}
where $s_0$ determines the closest possible surface point.
We first find the smallest $j$ for which $f_\theta$ changes from free space ($f_\theta < \tau$) to occupied space ($f_\theta\geq\tau$):
\begin{equation}
j = \argmin_{j'} \left(f_\theta(\bp_{j'+1}^\text{ray}) \geq \tau > f_\theta(\bp_{j'}^\text{ray})\right)
\end{equation}
We obtain an approximation to the surface depth $\hat d$ by applying the iterative secant method to the interval $[j \Delta s+s_0, (j+1)\Delta s+s_0]$.
In practice, we compute the surface depth for a batch of $N_p$ points in parallel.
It is important to note that we do not need to unroll the forward pass or store any intermediate results as we exploit implicit differentiation to directly obtain the gradient of $\hat{d}$ \wrt $\theta$.

\boldparagraph{Backward Pass}
The input to the backward pass is the gradient $\lambda = \tfrac{\partial \cL}{\partial \hat{d}}$ of the loss \wrt a single surface depth prediction.
The output of the backward pass is $\lambda \tfrac{\partial \hat d}{\partial \theta}$ , which can be computed using \eqref{eq:dhatd_dtheta}.
In practice, however, we would like to implement the backward pass not only for a single surface depth $\hat d$ but for a whole batch of depth values.

We can implement this efficiently by rewriting $\lambda \tfrac{\partial \hat d}{\partial \theta}$ as
\begin{equation}\label{eq:backward-efficient}
    \mu \frac{\partial \onet_\theta(\hat{\bp})}{\partial \theta}
    \quad \text{with} \quad
    \mu = 
    -\left(\frac{\partial \onet_\theta(\hat{\bp})}{\partial \hat{\bp}} \cdot \bw \right)^{-1}
    \lambda
\end{equation}
Importantly, the left term in \eqref{eq:backward-efficient}
corresponds to a normal backward operation applied to the neural network $\onet_\theta$ 
and the right term in \eqref{eq:backward-efficient} is just an (element-wise) scalar multiplication for all elements in the batch.
We can hence conveniently compute the backward pass of the operator $\theta \to \depthpred$ by first multiplying the incoming gradient $\lambda$ element-wise with a factor and then backpropagating the result through the operator $\theta \to \onet_\theta(\pointpred)$.
Both operations can be efficiently parallelized in common deep learning frameworks.

\subsection{Training}

During training, we assume that we are given $N$ images $\{\imggt_k\}_{k=1}^{N}$ together with corresponding camera intrinsics, extrinsics, and object masks $\{ \bM_k \}_{k=1}^N$.
As our experiments show, our method works with as little as one image per object. In addition, our method can also incorporate depth information $\{ \bD_k \}_{k=1}^N$, if available.

For training $\onet_\theta$ and $\texnet_\theta$, we randomly sample an image $\imggt_k$ and $N_p$ points $\bu$ on the image plane.
We distinguish the following three cases:
First, let $\cP_0$ denote the set of points $\bu$ that lie inside the object mask $\bM_k$ and for which the occupancy network predicts a finite surface depth $\hat d$ as described in~\secref{subsec:forward_pass}.
For these points we can define a loss $\cL_\text{rgb}(\theta)$ directly on the predicted image $\imgpred_k$.
Moreover, let $\cP_1$ denote the points $\bu$ which lie outside the object mask $\bM_k$.
While we cannot define a photometric loss for these points, we can define a loss $\cL_\text{freespace}(\theta)$ that encourages the network to remove spurious geometry along corresponding rays.
Finally, let $\cP_2$ denote the set of points $\bu$ which lie inside the object mask $\bM_k$, but for which the occupancy network does not predict a finite surface~depth~$\hat d$.
Again, we cannot use a photometric loss for these points, but we can define a loss $\cL_\text{occupancy}(\theta)$ that encourages the network to produce a finite surface depth.

\boldparagraph{RGB Loss}
For each point in $\cP_0$, we detect the predicted surface depth $\hat d$ as described in~\secref{subsec:forward_pass}.
We define a photo-consistency loss for the points as 
\begin{equation}
     \cL_\text{rgb}(\theta) 
     = \sum_{\bu \in \cP_0}
     \| \xi(\imggt)_\bu - \xi(\imgpred)_{\bu} \|
    \label{eq:loss_rgb}
\end{equation}
where $\xi(\cdot)$ computes image features and $\|\cdot\|$ defines a robust error metric.
In practice, we use RGB-values and (optionally) image gradients as features and an $\ell_1$-loss for $\|\cdot\|$.

\boldparagraph{Depth Loss}
When the depth is also given, we can directly incorporate an $\ell_1$ loss on the predicted surface depth as
\begin{equation}
    \cL_\text{depth}(\theta) 
    = \sum_{\bu \in \cP_0}
    | d - \hat d |_1
    \label{eq:loss_depth}
\end{equation}
where $d$ indicates the ground truth depth value of the sampled image point $\bu$ and $\hat d$ denotes the predicted surface depth for pixel $\bu$.

\boldparagraph{Freespace Loss} 
If a point $\bu$ lies outside the object mask but the predicted surface depth $\hat d$ is finite, the network falsely predicts surface point $\hat{\bp} = \br(\hat d)$.
Therefore, we penalize this occupancy with 
\begin{equation}
    \cL_\text{freespace}(\theta)
    = \sum_{\bu \in \cP_1}
    \text{BCE}(\onet_\theta(\hat{\bp}), 0)
\end{equation}
where $\text{BCE}$ is the binary cross entropy. 
When no surface depth is predicted, we apply the freespace loss to a randomly sampled point on the ray.

\boldparagraph{Occupancy Loss}
 If a point $\bu$ lies inside the object mask but the predicted surface depth $\hat d$ is infinite, the network falsely predicts no surface points on ray $\br$.
To encourage predicting occupied space on this ray, we uniformly sample depth values $d_\text{random}$ and define 
\begin{equation}
    \cL_\text{occupancy}(\theta) 
    = \sum_{\bu \in \cP_2}
    \text{BCE}(\onet_\theta(\br(d_\text{random})), 1)
\end{equation}
In the single-view reconstruction experiments, we instead use the first point on the ray which lies inside all object masks (depth of the visual hull).
If we have additional depth supervision, we use the ground truth depth for the occupancy loss.
Intuitively, $\cL_\text{occupancy}$ encourages the network to occupy space along the respective rays which can then be used by $\cL_\text{rgb}$ in~\eqref{eq:loss_rgb} and $\cL_\text{depth}$ in~\eqref{eq:loss_depth} to refine the initial occupancy.

\boldparagraph{Normal Loss}
Optionally, our representation allows us to incorporate a smoothness prior by regularizing surface normals.
This is useful especially for real-world data as training with 2D or 2.5D supervision includes unconstrained areas where this prior enforces more natural shapes.
We define this loss as 
\begin{equation}
    \cL_{\text{normal}}(\theta) = \sum_{\bu \in \cP_0} \left|\left| \bn(\hat{\bp_\bu}) - \bn(\bq_\bu) \right|\right|_2
\end{equation}
where $\bn(\cdot)$ denotes the normal vector, $\hat{\bp_\bu}$ the predicted surface point and $\bq_\bu$ a randomly sampled neighbor of $\hat{\bp_\bu}$.\footnote{See supplementary for details.}

\subsection{Implementation Details}

We implement the combined network with $5$ fully-connected ResNet \cite{He2016CVPR} blocks and ReLU activation.
The output dimension of the last layer is $4$, one dimension for the occupancy probability and three dimensions for the texture.
For the single-view reconstruction experiments, we encode the input image with an ResNet-18~\cite{He2016CVPR} encoder network $\bg_\phi$ which outputs a $256$-dimensional latent code $\text{z}$.
To facilitate training, we start with a ray sampling accuracy of $n=16$ which we iteratively increase to $n=128$ by doubling $n$ after $50$, $150$, and $250$ thousand iterations.
We choose the sampling interval $[s_0,n\Delta s+s_0]$ such that it covers the volume of interest for each object.
We set $\tau = 0.5$ for all experiments.
We train on a single NVIDIA~V100 GPU with a batch size of $64$ images with $1024$ random pixels each.
We use the Adam optimizer~\cite{Kingma2015ICLR} with learning rate $\gamma = 10^{-4}$ which we decrease by a factor of $5$ after $750$ and $1000$ epochs, respectively.
\section{Experiments}

We conduct two different types of experiments to validate our approach. 
First, we investigate how well our approach reconstructs 3D shape and texture from a \textbf{single RGB image} when trained on a large collection of RGB or RGB-D images.
Here, we consider both the case where we have access to multi-view supervision
and the case where we use only a single RGB-D image per object during training.
Next, we apply our approach to the challenging task of \textbf{multi-view reconstruction}, where the goal is to
reconstruct complex 3D objects from real-world multi-view imagery. %

\subsection{Single-View Reconstruction}

\label{subsec:single-view-reconstruction}
First, we investigate to which degree our method can infer a 3D shape and texture representation from single-views. %
We train a single model jointly on all categories.

\boldparagraph{Datasets} %
To adhere to community standards~\cite{Mescheder2019CVPR, Wang2018ECCVc,Choy2016ECCV}, we use the Choy~\etal~\cite{Choy2016ECCV} subset ($13$ classes) of the ShapeNet~dataset~\cite{Chang2015ARXIV} for 2.5D and 3D supervised methods with training, validation, and test splits from \cite{Mescheder2019CVPR}.
While we use the renderings from Choy~\etal~\cite{Choy2016ECCV} as input, we additionally render $24$ images of resolution $256^2$ with depth maps and object masks per object which we use for supervision. 
We randomly sample the viewpoint on the northern hemisphere as well as the distance of the camera to the object to get diverse supervision data. %
For 2D supervised methods, we adhere to community standards~\cite{LIU2019ICCV, KATO2018CVPR, Yan2016NIPS} and use the renderings and splits from~\cite{KATO2018CVPR}.
Similar to~\cite{Mescheder2019CVPR,Choy2016ECCV,KATO2018CVPR}, we train with objects in canonical pose.

\boldparagraph{Baselines} %
We compare against the following methods which all produce watertight meshes as output:
3D-R2N2~\cite{Choy2016ECCV} (voxel-based), Pixel2Mesh~\cite{Wang2018ECCVc} (mesh-based), and ONet~\cite{Mescheder2019CVPR} (implicit representation).
We further compare against both the 2D and the 2.5D supervised version of Differentiable Ray Consistency (DRC) \cite{Tulsiani2017CVPR} (voxel-based) and the 2D supervised Soft Rasterizer (SoftRas) \cite{LIU2019ICCV} (mesh-based).
For 3D-R2N2, we use the pre-trained model from \cite{Mescheder2019CVPR} which was shown to produce better results than the original model from \cite{Choy2016ECCV}.
For the other baselines, we use the pre-trained models\footnote{%
Unfortunately, we cannot show texture results for DRC and SoftRas as texture prediction is not part of the official code repositories.
}
from the authors.

\subsubsection{Multi-View Supervision}
We first consider the case where we have access to multi-view supervision with $N=24$ images and corresponding object masks.
In addition, we also investigate the case when ground truth depth maps are given.

\boldparagraph{Results}
We evaluate the results using the Chamfer-$L_1$ distance from \cite{Mescheder2019CVPR}. In contrast to previous works \cite{Mescheder2019CVPR,LIU2019ICCV,Tulsiani2017CVPR,Choy2016ECCV}, we compare directly \wrt to the ground truth shape models, not the voxelized or watertight versions. 
\begin{table*}
  \centering
  \resizebox{1.\linewidth}{!}{   
  \begin{tabular}{l|ccc|cc|ccc}
\toprule
{} & \multicolumn{3}{c|}{2D Supervision} & \multicolumn{2}{c|}{2.5D Supervision} & \multicolumn{3}{c}{3D Supervision} \\
{} & DRC (Mask)~\cite{Tulsiani2017CVPR} & SoftRas~\cite{LIU2019ICCV} & Ours ($\cL_\text{RGB}$) & DRC (Depth)~\cite{Tulsiani2017CVPR} & Ours ($\cL_\text{Depth}$) & 3D R2N2~\cite{Choy2016ECCV} & ONet~\cite{Mescheder2019CVPR} & Pixel2Mesh~\cite{Wang2018ECCVc} \\
category    &                                    &                            &                         &                                     &                           &                             &                               &                                 \\
\midrule
airplane    &                              0.659 &             \textbf{0.149} &                   0.190 &                               0.377 &            \textbf{0.143} &                       0.215 &                \textbf{0.151} &                           0.183 \\
bench       &                                  - &                      0.241 &          \textbf{0.210} &                                   - &            \textbf{0.165} &                       0.210 &                \textbf{0.171} &                           0.191 \\
cabinet     &                                  - &                      0.231 &          \textbf{0.220} &                                   - &            \textbf{0.183} &                       0.246 &                \textbf{0.189} &                           0.194 \\
car         &                              0.340 &                      0.221 &          \textbf{0.196} &                               0.316 &            \textbf{0.179} &                       0.250 &                         0.181 &                  \textbf{0.154} \\
chair       &                              0.660 &                      0.338 &          \textbf{0.264} &                               0.510 &            \textbf{0.226} &                       0.282 &                \textbf{0.224} &                           0.259 \\
display     &                                  - &                      0.284 &          \textbf{0.255} &                                   - &            \textbf{0.246} &                       0.323 &                         0.275 &                  \textbf{0.231} \\
lamp        &                                  - &             \textbf{0.381} &                   0.413 &                                   - &            \textbf{0.362} &                       0.566 &                         0.380 &                  \textbf{0.309} \\
loudspeaker &                                  - &                      0.320 &          \textbf{0.289} &                                   - &            \textbf{0.295} &                       0.333 &                         0.290 &                  \textbf{0.284} \\
rifle       &                                  - &             \textbf{0.155} &                   0.175 &                                   - &            \textbf{0.143} &                       0.199 &                         0.160 &                  \textbf{0.151} \\
sofa        &                                  - &                      0.407 &          \textbf{0.224} &                                   - &            \textbf{0.221} &                       0.264 &                         0.217 &                  \textbf{0.211} \\
table       &                                  - &                      0.374 &          \textbf{0.280} &                                   - &            \textbf{0.180} &                       0.247 &                \textbf{0.185} &                           0.215 \\
telephone   &                                  - &             \textbf{0.131} &                   0.148 &                                   - &            \textbf{0.130} &                       0.221 &                         0.155 &                  \textbf{0.145} \\
vessel      &                                  - &             \textbf{0.233} &                   0.245 &                                   - &            \textbf{0.206} &                       0.248 &                         0.220 &                  \textbf{0.201} \\
\midrule
mean        &                              0.553 &                      0.266 &          \textbf{0.239} &                               0.401 &            \textbf{0.206} &                       0.277 &                         0.215 &                  \textbf{0.210} \\
\bottomrule
\end{tabular}  
  }
  \caption{
    \textbf{Single-View Reconstruction.} 
    We report Chamfer-$L_1$ distances \wrt the ground truth meshes for the single-view experiment.
    We compare against Differentiable Ray Consistency (DRC) \cite{Tulsiani2017CVPR} (2D and 2.5D supervision), Soft Rasterizer \cite{LIU2019ICCV} (2D supervision), 3D-R2N2 \cite{Choy2016ECCV}, Occupancy Networks (ONet) \cite{Mescheder2019CVPR}, and Pixel2Mesh \cite{Wang2018ECCVc} (all 3D supervision).
  }
  \label{tab:single-view-reconstruction}
  \vspace{-.35cm}
\end{table*}
\begin{figure}
  \centering  %
    \setlength\tabcolsep{4pt} %
    \begin{tabular}{p{1cm}cccc}
       {\small Input} & {\small SoftRas} & {\small Ours ($\cL_\text{RGB}$)} & {\small Pixel2Mesh} & {\small Ours ($\cL_\text{Depth}$)} \\
      \midrule
      \includegraphics[width=0.75\linewidth, trim={.2cm, .2cm, .2cm, .2cm}, clip]{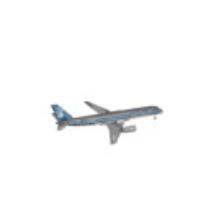} &
      \includegraphics[width=0.12\linewidth, trim={2.3cm, 2.cm, 2.3cm, 3cm}, clip]{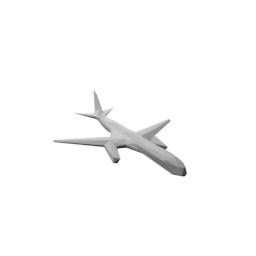} &
      \includegraphics[width=0.12\linewidth, trim={2.3cm, 2.cm, 2.3cm, 3cm}, clip]{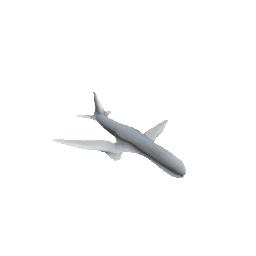} &
      \includegraphics[width=0.12\linewidth, trim={2.3cm, 2.cm, 2.3cm, 3cm}, clip]{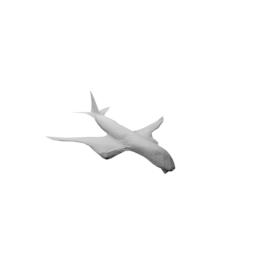} & 
      \includegraphics[width=0.12\linewidth, trim={2.3cm, 2.cm, 2.3cm, 3cm}, clip]{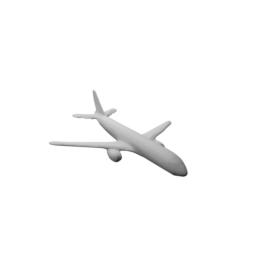} \\
      \includegraphics[width=0.75\linewidth, trim={.0cm, .0cm, .0cm, .0cm}, clip]{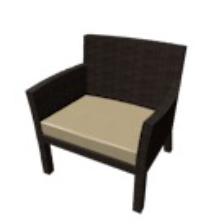} &
      \includegraphics[width=0.12\linewidth, trim={.5cm, .0cm, .3cm, .5cm}, clip]{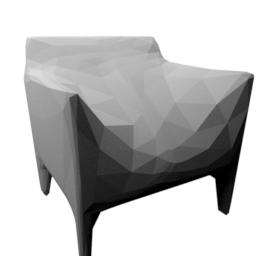} &
      \includegraphics[width=0.12\linewidth, trim={1cm, .5cm, 1cm, 1cm}, clip]{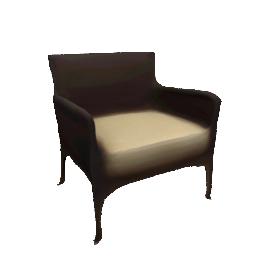} &
      \includegraphics[width=0.12\linewidth, trim={.5cm, .0cm, .3cm, .5cm}, clip]{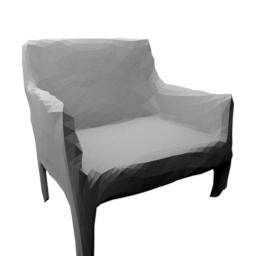} & 
      \includegraphics[width=0.12\linewidth, trim={.5cm, .0cm, .3cm, .5cm}, clip]{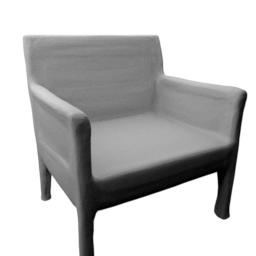} \\
      \includegraphics[width=0.75\linewidth, trim={.0cm, .0cm, .0cm, .0cm}, clip]{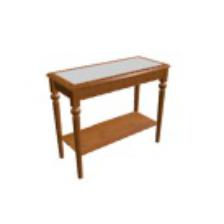} &
      \includegraphics[width=0.12\linewidth, trim={0cm, .5cm, 0cm, 2.cm}, clip]{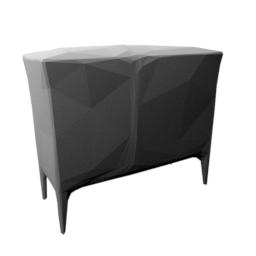} &
      \includegraphics[width=0.12\linewidth, trim={0cm, .5cm, 0cm, 2.cm}, clip]{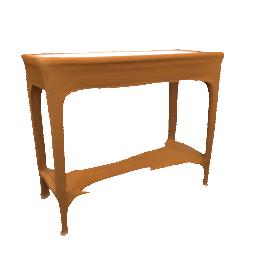} &
      \includegraphics[width=0.12\linewidth, trim={0cm, .5cm, 0cm, 2.cm}, clip]{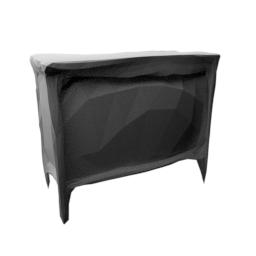} & 
      \includegraphics[width=0.12\linewidth, trim={0cm, .5cm, 0cm, 2.cm}, clip]{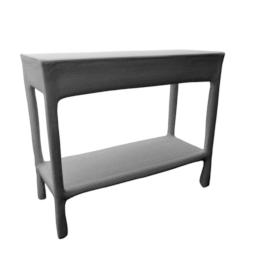} \\
      \includegraphics[width=0.75\linewidth, trim={.0cm, .0cm, .0cm, .0cm}, clip]{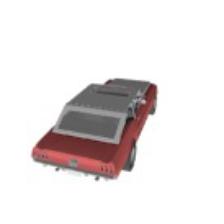} &
      \includegraphics[width=0.12\linewidth, trim={1.5cm, 2.cm, 1.5cm, 3.0cm}, clip]{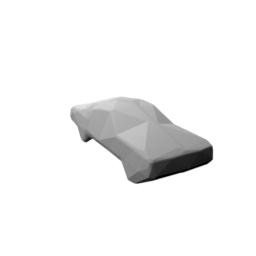} &
      \includegraphics[width=0.12\linewidth, trim={1.5cm, 2.cm, 1.5cm, 3.0cm}, clip]{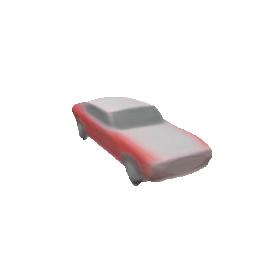} &
      \includegraphics[width=0.12\linewidth, trim={1.5cm, 2.cm, 1.5cm, 3.0cm}, clip]{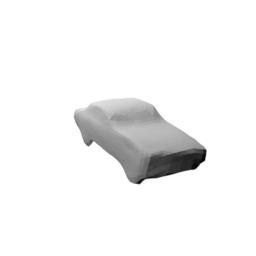} & 
      \includegraphics[width=0.12\linewidth, trim={1.5cm, 2.cm, 1.5cm, 3.0cm}, clip]{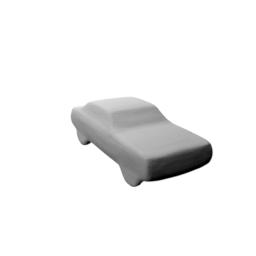} \\
      \includegraphics[width=0.75\linewidth, trim={.0cm, .5cm, .0cm, .0cm}, clip]{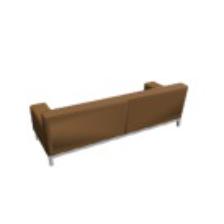} &
      \includegraphics[width=0.12\linewidth, trim={1cm 2.2cm 1.5cm 3cm}, clip]{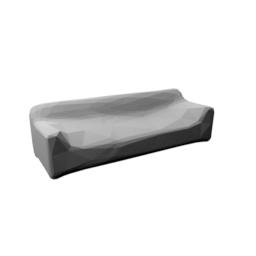} &
      \includegraphics[width=0.12\linewidth, trim={1cm 2.2cm 1.5cm 3cm}, clip]{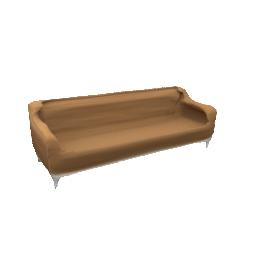} &
      \includegraphics[width=0.12\linewidth, trim={1cm 2.2cm 1.5cm 3cm}, clip]{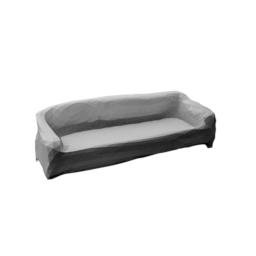} & 
      \includegraphics[width=0.12\linewidth, trim={1cm 2.2cm 1.5cm 3cm}, clip]{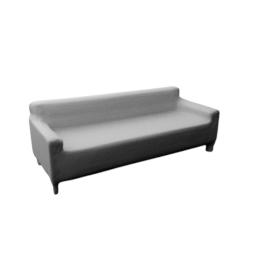} \\
  \end{tabular} 
  \caption{
    \textbf{Single-View Reconstruction.}
    We show the input renderings from~\cite{Choy2016ECCV} and the output of
    our 2D supervised ($\cL_\text{RGB}$) and 2.5D supervised ($\cL_\text{Depth}$) model, Soft Rasterizer~\cite{LIU2019ICCV} and Pixel2Mesh~\cite{Wang2018ECCVc}. 
    For 2D supervised methods we use a corresponding view from~\cite{KATO2018CVPR} as input.
  }
  \label{tab:single-view-reconstruction-img}
  \vspace{-.2cm}
\end{figure}

In~\tabref{tab:single-view-reconstruction} and \figref{tab:single-view-reconstruction-img} we show quantitative and qualitative results for our method and various baselines.
We can see that our method is able to infer accurate 3D shape and texture representations from single-view images when only trained on multi-view images and object masks as supervision signal.
Quantitatively (\tabref{tab:single-view-reconstruction}), our method performs best among the approaches with 2D supervision and rivals the quality of methods with full 3D supervision.
When trained with depth, our method performs comparably to the methods which use full 3D information. 
Qualitatively (\figref{tab:single-view-reconstruction-img}), we see that in contrast to the mesh-based approaches, our method is not restricted to certain topologies.
When trained with the photo-consistency loss $\cL_\text{RGB}$, we see that our approach is able to predict accurate texture information in addition to the 3D shape.

\subsubsection{Single-View Supervision}
\begin{figure}
  \centering
    \setlength\tabcolsep{4pt} %
    \begin{tabular}{C{.7cm}C{1.4cm}}
      {\small Input} & {\small Prediction} \\
     \midrule
     \includegraphics[width=1.\linewidth,  clip]{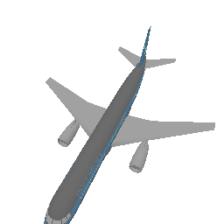} &
     \includegraphics[width=1.\linewidth, trim={1cm 1cm 1cm 1cm}, clip]{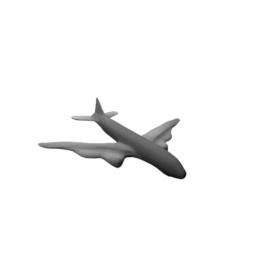} \\
     \includegraphics[width=1.\linewidth, clip]{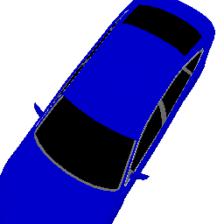} &
     \includegraphics[width=1.\linewidth, trim={1cm 2.1cm 1cm 2.5cm}, clip]{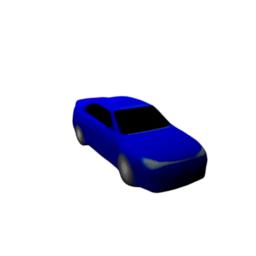}
 \end{tabular}
 \hspace{1cm}
    \begin{tabular}{C{.7cm}C{1.4cm}}
       {\small Input} & {\small Prediction} \\
      \midrule
      \includegraphics[width=1.\linewidth,  clip]{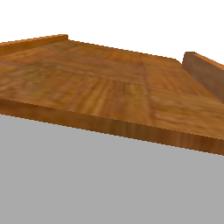} &
      \includegraphics[width=1.\linewidth, trim={1cm 1cm 1.cm 1cm}, clip]{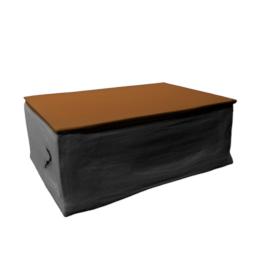} \\
      \includegraphics[width=1.\linewidth,  clip]{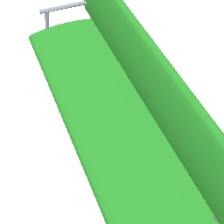} &
      \includegraphics[width=1.\linewidth, trim={1cm 2.1cm 1cm 2.5cm}, clip]{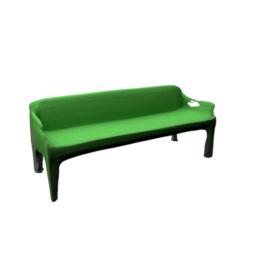}
  \end{tabular}
      \caption{
        \textbf{Single-View Reconstruction with Single-View Supervision.}
        While only trained with a single-view per object, our model predicts accurate 3D geometry and texture. %
      }
      \label{fig:single-image-single-view}
      \vspace{-.2cm}
\end{figure}

The previous experiment indicates that our model is able to infer accurate shape and texture information without 3D supervision.  %
A natural question to ask is how many images are required during training.
To this end, we investigate the case when only \textit{a single image} with depth and camera information is available.
Since we represent the 3D shape in a canonical object coordinate system, the hypothesis is that the model can aggregate the information over multiple training instances, although it sees every object only from one perspective.
As the same image is used both as input and supervision signal, we now condition on our renderings instead of the ones provided by Choy~\etal~\cite{Choy2016ECCV}.

\boldparagraph{Results}
Surprisingly, \figref{fig:single-image-single-view} shows that our method can infer appropriate 3D shape and texture when only a single-view is available per object, confirming our hypothesis.
Quantitatively, the Chamfer distance of the model trained with $\cL_\text{RGB}$ and $\cL_\text{Depth}$ with only a single view ($0.410$) is comparable to the model trained with $\cL_\text{Depth}$ with $24$ views ($0.383$).
The reason for the numbers being worse than in~\secref{subsec:single-view-reconstruction} is that for our renderings, we do not only sample the viewpoint, but also the distance to the object resulting in a much harder task (see \figref{fig:single-image-single-view}).

\subsection{Multi-View Reconstruction}
\begin{figure*}
  \centering
    \begin{subfigure}{.32\linewidth}
      \includegraphics[width=1.\linewidth, trim={12cm, 17.5cm, 15cm, 16cm}, clip]{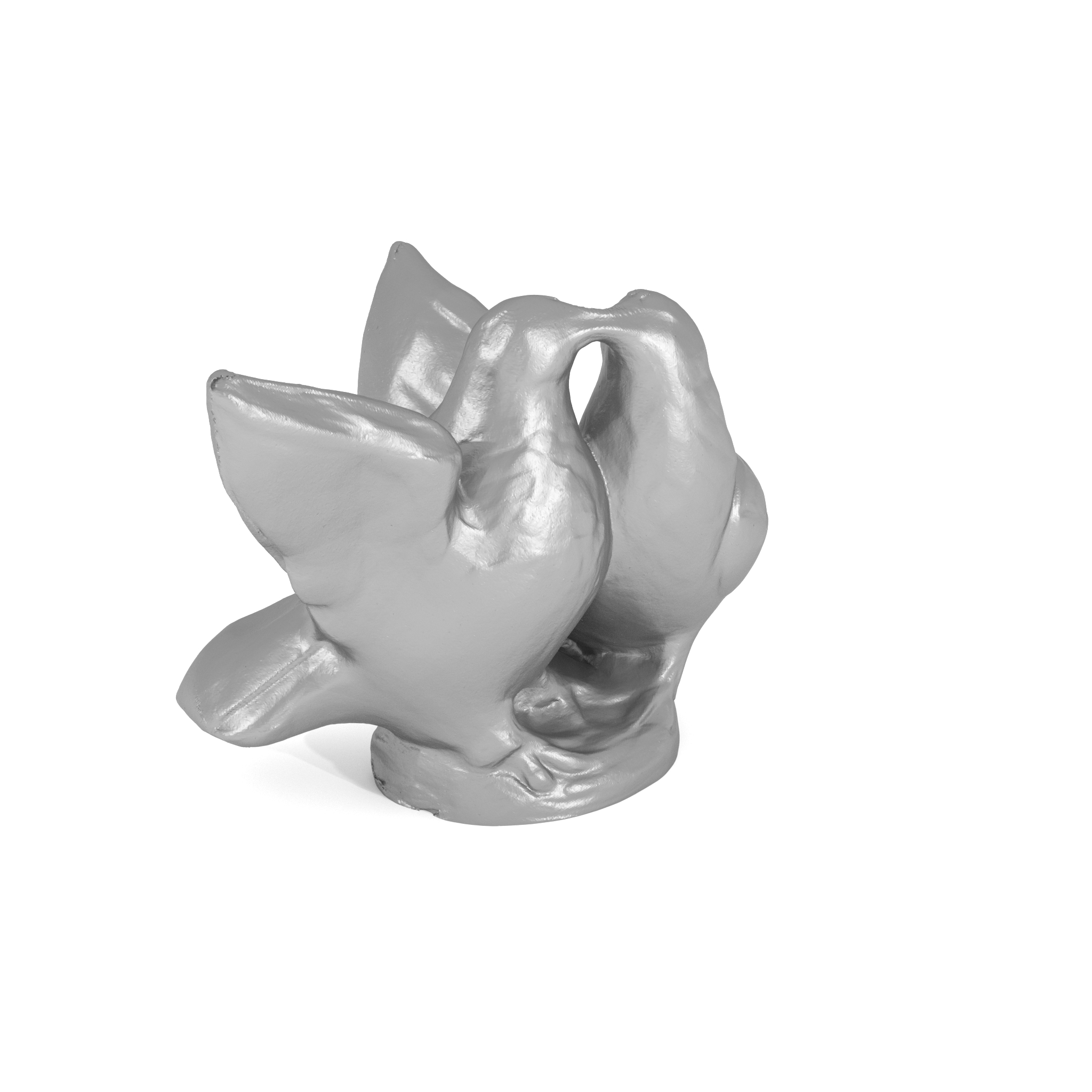}
      \caption{Shape}
    \end{subfigure} 
    \begin{subfigure}{.32\linewidth}
      \includegraphics[width=1.\linewidth, trim={12cm, 17.5cm, 15cm, 16cm}, clip]{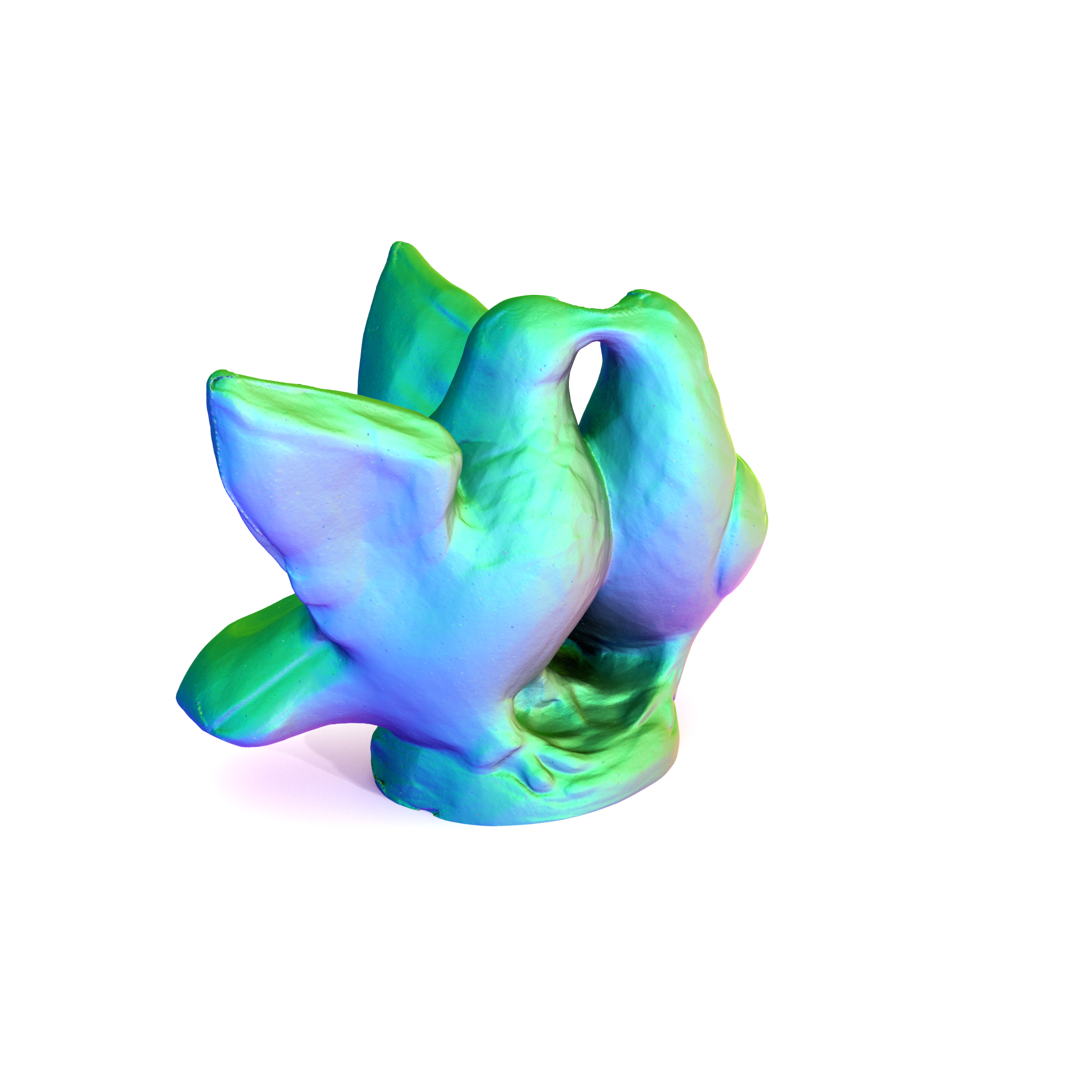}
      \caption{Normals}
    \end{subfigure}
    \begin{subfigure}{.32\linewidth}
      \includegraphics[width=1.\linewidth, trim={12cm, 17.5cm, 15cm, 16cm}, clip]{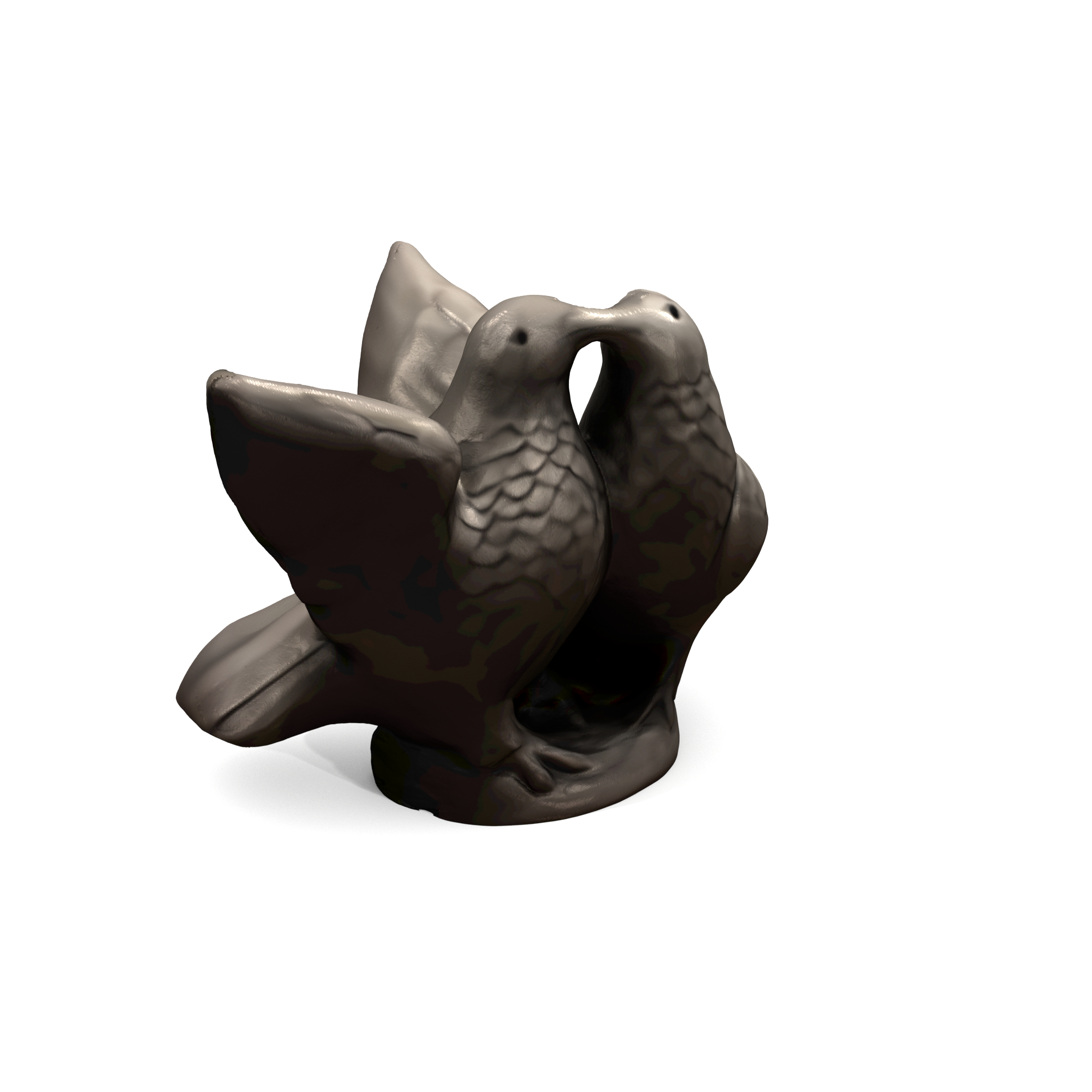}
      \caption{Texture}
    \end{subfigure}
  \caption{
    \textbf{Multi-View Stereo.}
    We show the shape, normals, and the textured shape for our method trained with 2D images and sparse depth maps for scan $106$ of the DTU~dataset~\cite{Aanes2016IJCV}.
    \vspace{-.35cm}
    }
\label{fig:mvs}
\end{figure*}
\begin{figure}
  \centering
  \begin{subfigure}{.32\linewidth}
    \includegraphics[width=1.\linewidth, trim={8cm, 8cm, 8cm, 0cm}, clip]{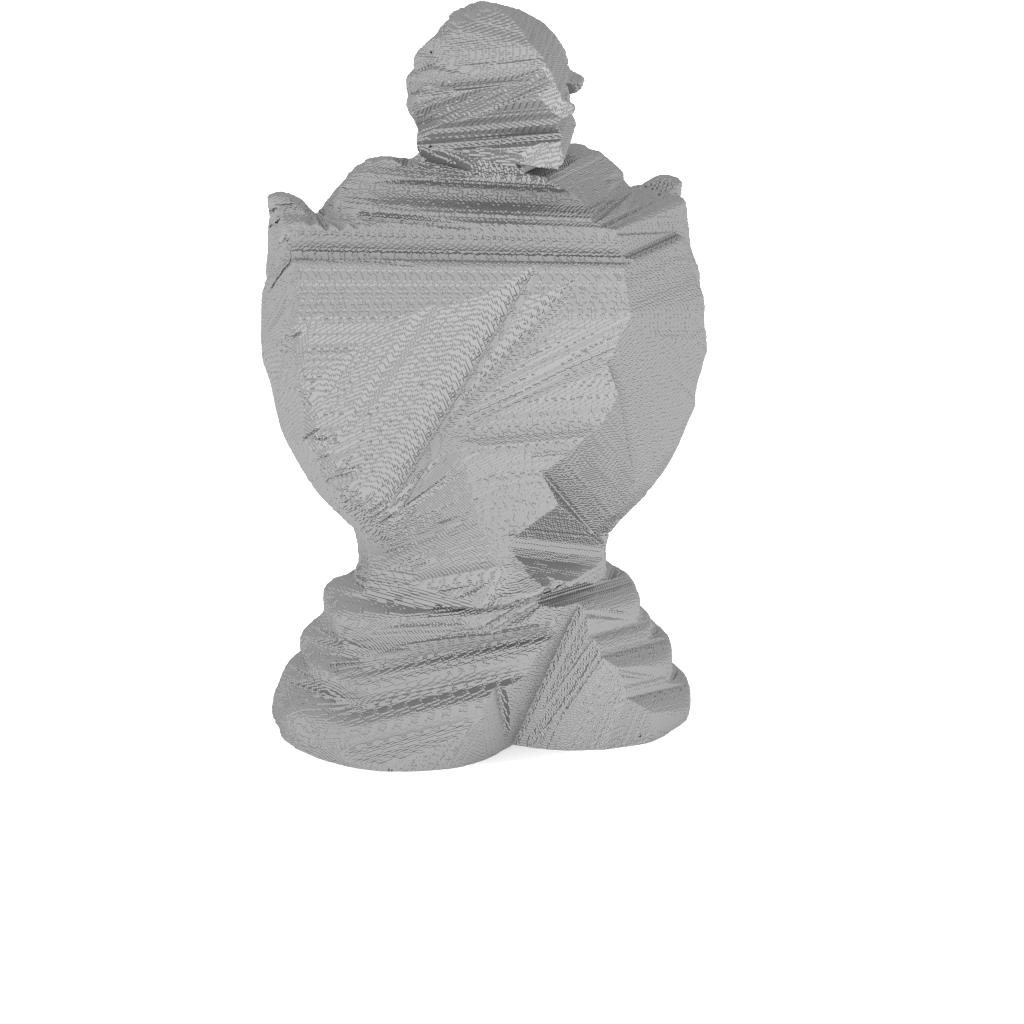}
    \caption{Visual Hull \cite{Laurentini1994PAMI}}
  \end{subfigure}
  \begin{subfigure}{.32\linewidth}
    \includegraphics[width=1.\linewidth, trim={9.5cm, 8cm, 9cm, 4cm}, clip]{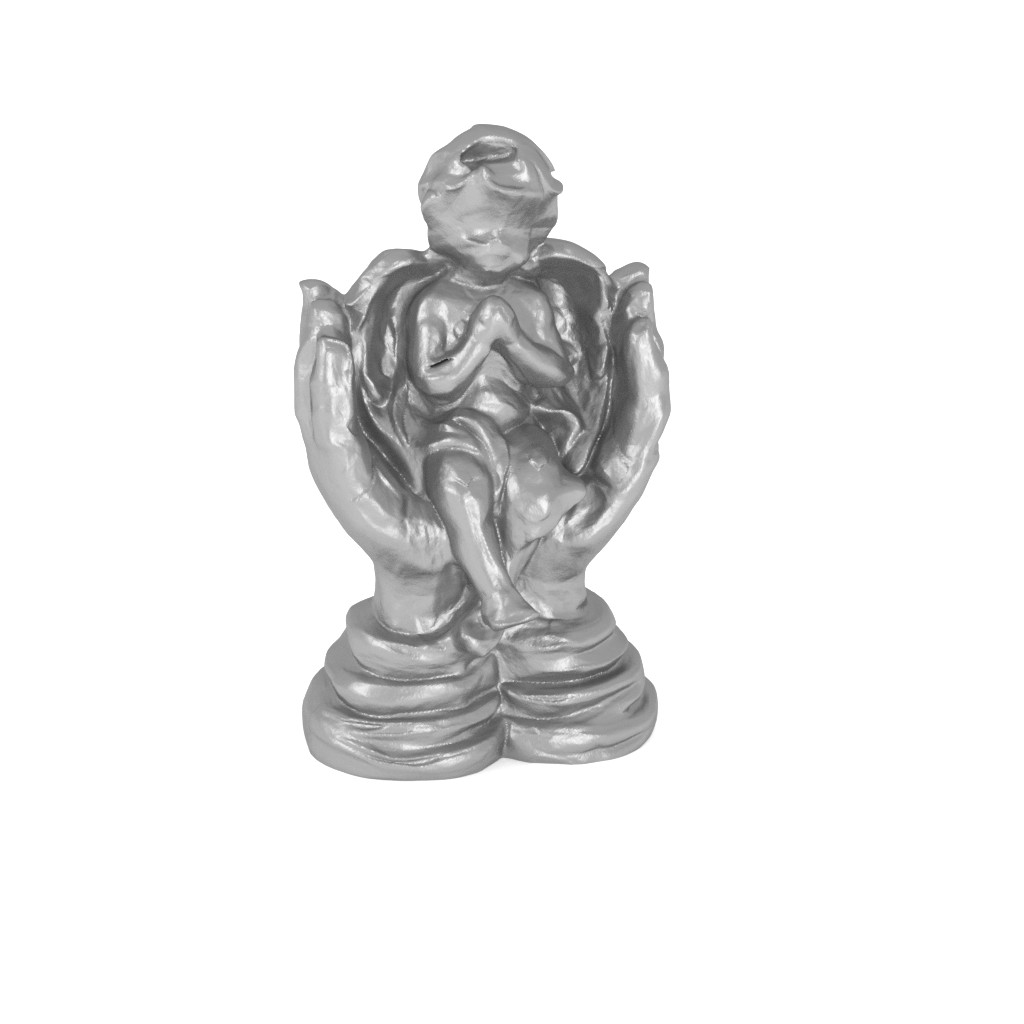}
    \caption{Ours ($\cL_\text{RGB}$)}
  \end{subfigure}
  \begin{subfigure}{.32\linewidth}
    \includegraphics[width=1.\linewidth, trim={8cm, 8cm, 8cm, 0cm}, clip]{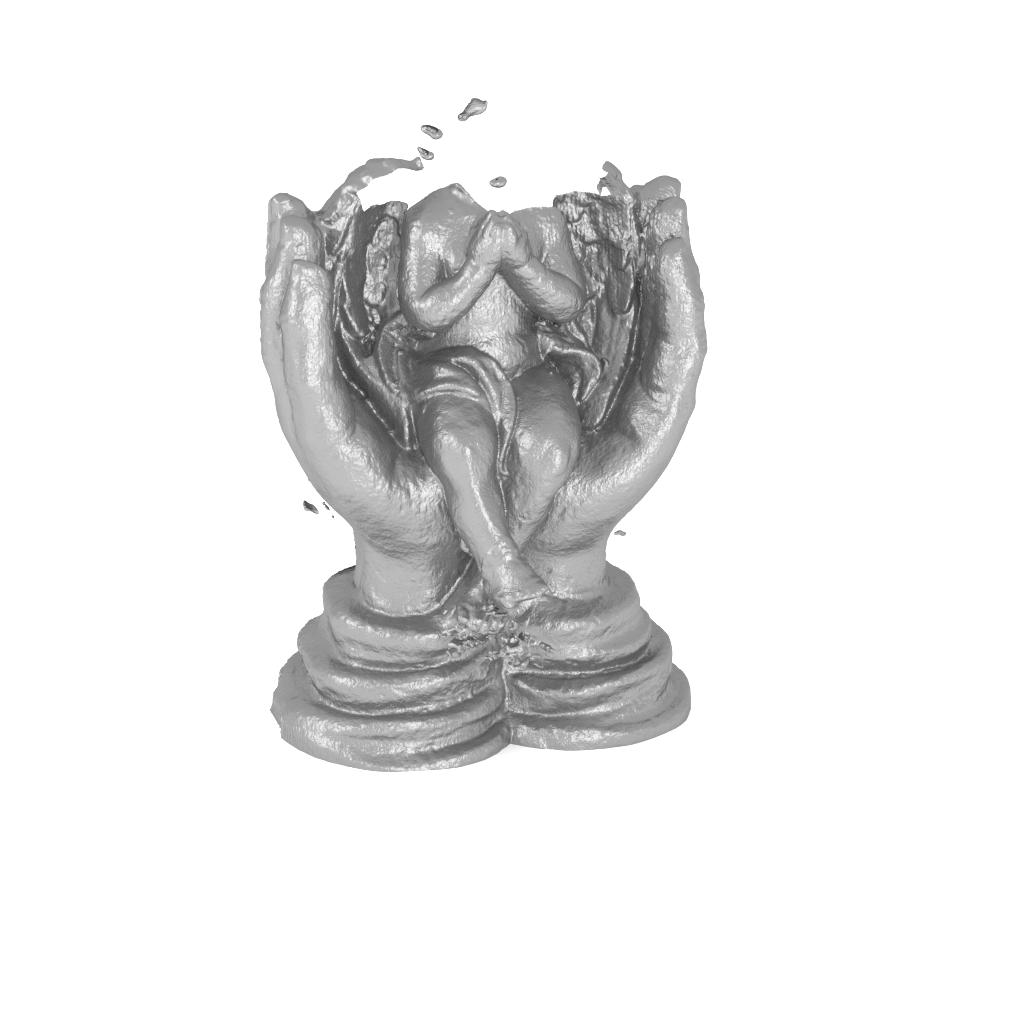}
    \caption{Ground Truth}
    \label{fig:mvs-visual-hull-gt} 
  \end{subfigure}
  \caption{
    \textbf{Comparison against Visual Hull.}
    We show the visual hull, the shape prediction of our model trained with $\cL_\text{RGB}$, and the ground truth for scan~$118$ of the DTU~dataset.
    Our method uses RGB cues to improve over the visual hull and predicts parts that are missing in the ground truth.
  }
\vspace{-.2cm} 
\label{fig:mvs-visual-hull} 
\end{figure}
Finally, we investigate if our method is also applicable to multi-view reconstruction in real-world scenarios.
We investigate two cases: First, when multi-view images and object masks are given. 
Second, when additional sparse depth maps are given which can be obtained from classic multi-view stereo algorithms~\cite{Schonberger2016CVPR}.
For this experiment, we do not condition our model and train one model per object.

\boldparagraph{Dataset}
We conduct this experiment on scans $65$, $106$, and $118$ from the challenging real-world DTU~dataset~\cite{Aanes2016IJCV}.
The dataset contains $49$ or $65$ images with camera information for each object and baseline and structured light ground truth data.
The presented objects are challenging as their appearance changes in different viewpoints due to specularities.
Our sampling-based approach allows us to train on the full image resolution of $1200 \times 1600$.
We label the object masks ourselves and always remove the same images with profound changes in lighting conditions, \eg, caused by the appearance of scanner parts in the background. %

\boldparagraph{Baselines}
We compare against classical approaches that have 3D meshes as output.
To this end, we run screened Poisson surface reconstruction~(sPSR)~\cite{Kazhdan2013SIGGRAPH} on the output of the classical MVS algorithms Campbell~\etal~\cite{Campbell2008ECCV}, Furukawa~\etal~\cite{Furukawa2009PAMI}, Tola~\etal~\cite{Tola2011MVA}, and Colmap~\cite{Schonberger2016CVPR}.
We find that the results on the DTU benchmark for the baselines are highly sensitive to the trim parameter of sPSR and therefore report results for the trim parameters $0$ (watertight output), $5$ (good qualitative results) and $7$ (good quantitative results). 
For a fair comparison, we use the object masks to remove all points which lie outside the visual hull from the predictions of the baselines before running sPSR.\footnote{See supplementary material for details.}
We use the official DTU evaluation script in ``surface mode''.

\begin{table}
    \centering
    \resizebox{1.\linewidth}{!}{   
    \begin{tabular}{lcccc}
\toprule
{} &  Trim Param. &        Accuracy &    Completeness &   Chamfer-$L_1$ \\
\midrule
Tola~\cite{Tola2011MVA} + sPSR           &            0 &           2.409 &           1.242 &           1.826 \\
Furu~\cite{Furukawa2009PAMI} + sPSR      &            0 &           2.146 &           0.888 &           1.517 \\
Colmap~\cite{Schonberger2016CVPR} + sPSR &            0 &  \textbf{1.881} &           0.726 &  \textbf{1.303} \\
Camp~\cite{Campbell2008ECCV} + sPSR      &            0 &           2.213 &  \textbf{0.670} &           1.441 \\
\midrule
Tola~\cite{Tola2011MVA} + sPSR           &            5 &           1.531 &           1.267 &           1.399 \\
Furu~\cite{Furukawa2009PAMI} + sPSR      &            5 &           1.733 &           0.888 &           1.311 \\
Colmap~\cite{Schonberger2016CVPR} + sPSR &            5 &  \textbf{1.400} &           0.782 &  \textbf{1.091} \\
Camp~\cite{Campbell2008ECCV} + sPSR      &            5 &           1.991 &  \textbf{0.670} &           1.331 \\
\midrule
Tola~\cite{Tola2011MVA} + sPSR           &            7 &  \textbf{0.396} &           1.424 &           0.910 \\
Furu~\cite{Furukawa2009PAMI} + sPSR      &            7 &           0.723 &           0.955 &           0.839 \\
Colmap~\cite{Schonberger2016CVPR} + sPSR &            7 &           0.446 &           1.020 &  \textbf{0.733} \\
Camp~\cite{Campbell2008ECCV} + sPSR      &            7 &           1.466 &  \textbf{0.719} &           1.092 \\
\midrule
Ours ($\cL_\text{RGB}$)                      &           - &           1.054 &  \textbf{0.760} &           0.907 \\
Ours ($\cL_\text{RGB}$ + $\cL_\text{Depth}$) &           - &  \textbf{0.789} &           0.775 &  \textbf{0.782} \\
\bottomrule
\end{tabular}
  
    }
    \caption{
      \textbf{Multi-View Stereo.}
      We show quantitative results for scans $65$, $106$, and $118$ on the DTU~dataset.
      For the baselines, we perform screened Poisson surface reconstruction (sPSR)~\cite{Kazhdan2013SIGGRAPH} with trim parameters $0$, $5$, and $7$ to obtain the final output.
      It shows that our generic method achieves results comparable to the highly optimized MVS methods.
      }
    \label{tab:mvs}
\end{table}

\begin{figure}
  \centering
  \begin{subfigure}{.32\linewidth}
    \includegraphics[width=1.\linewidth, trim={3cm 9.5cm 3cm 5cm}, clip]{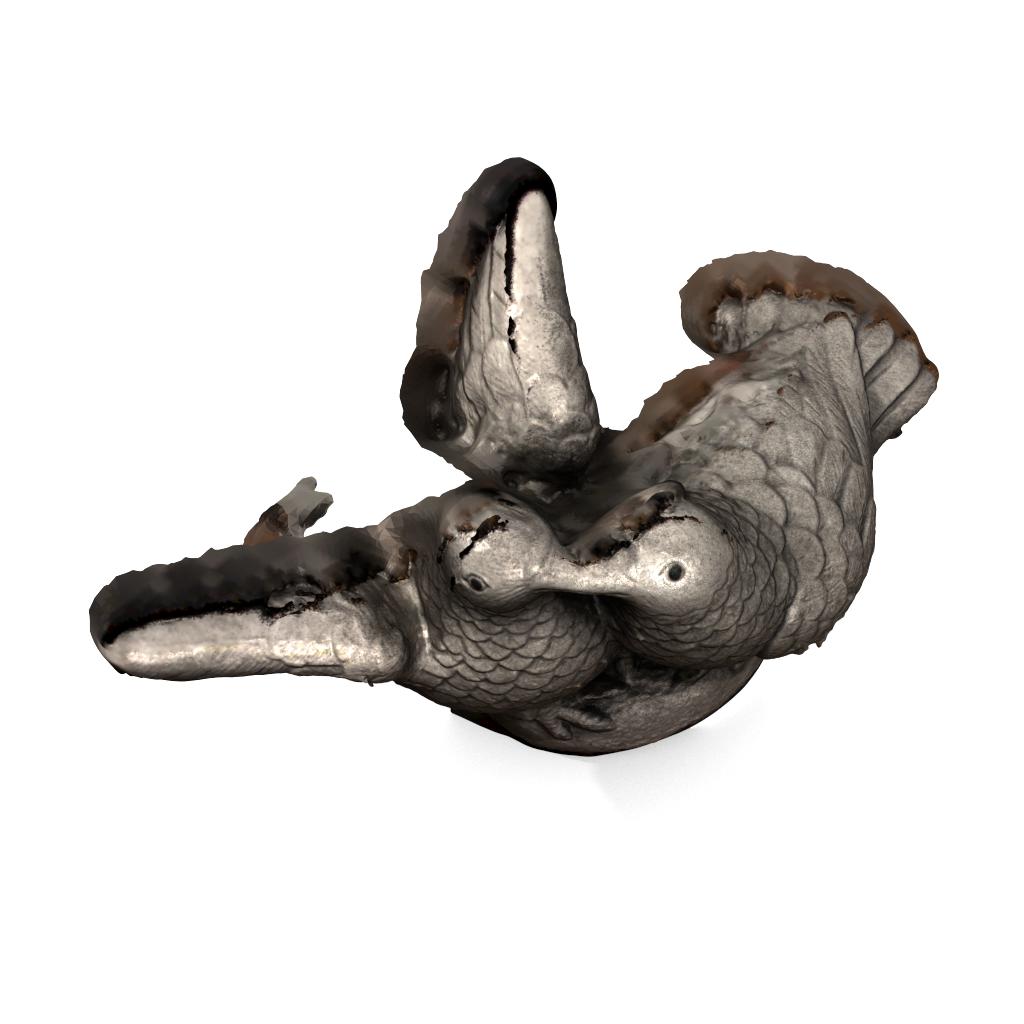}
    \caption{Colmap $5$}
    \label{subfig:colmap5}
  \end{subfigure}
  \begin{subfigure}{.32\linewidth}
    \includegraphics[width=1.\linewidth, trim={3cm 9.5cm 3cm 5cm}, clip]{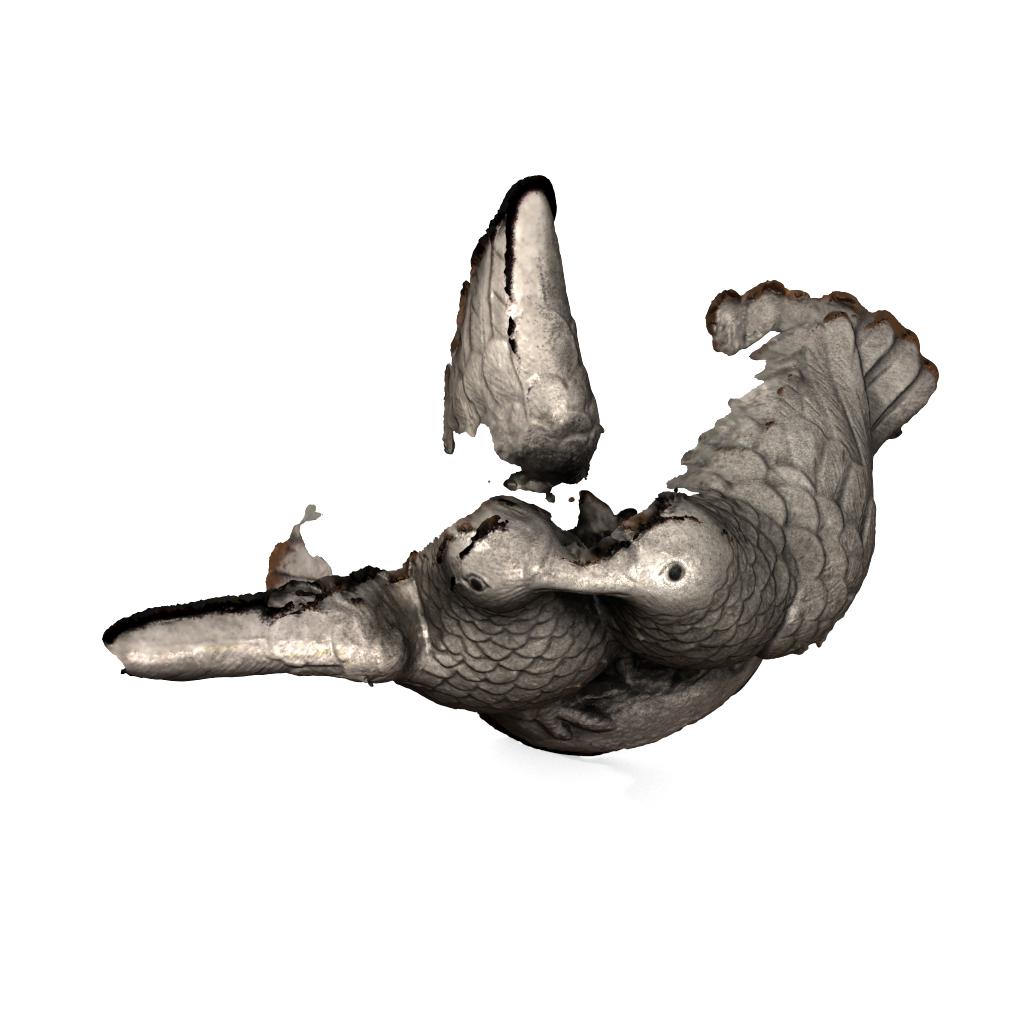}
    \caption{Colmap $7$}
    \label{subfig:colmap7}
  \end{subfigure}
  \begin{subfigure}{.32\linewidth}
    \includegraphics[width=1.\linewidth, trim={3cm 9.5cm 3cm 5cm}, clip]{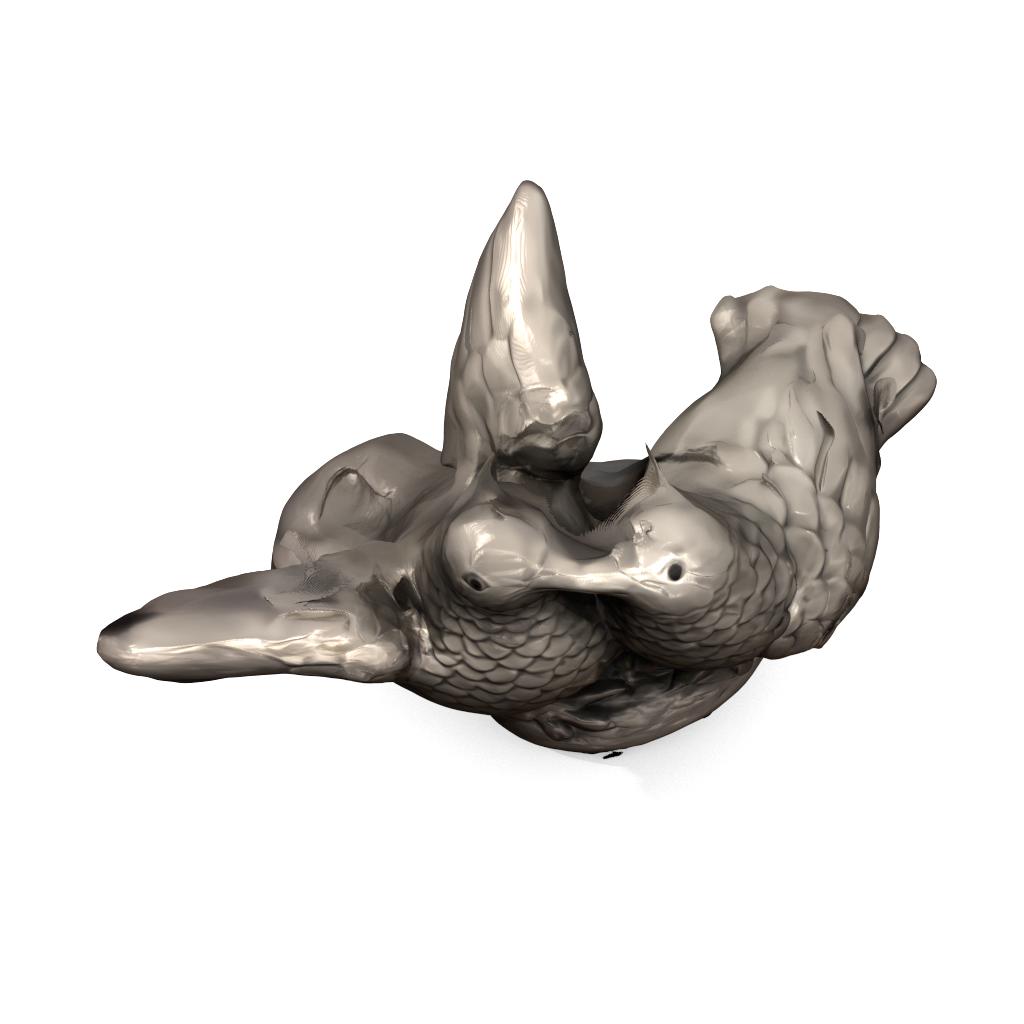}
    \caption{Ours}
  \end{subfigure}
  \caption{
    \textbf{Effect of Trim Parameter.}
    We show screened Poisson surface reconstructions~\cite{Kazhdan2013SIGGRAPH} with trim parameters $5$ and $7$ for Colmap~\cite{Schonberger2016CVPR} and the prediction of our model trained with $\cL_\text{RGB} + \cL_\text{Depth}$ for scan~$106$ of the DTU~dataset.
  }
  \vspace{-.2cm}
\label{fig:trim-parameter} 
\end{figure}

\boldparagraph{Results}
We show qualitative and quantitative results in~\figref{fig:mvs} and \tabref{tab:mvs}.
Qualitatively, we find that our method can be used for multi-view 3D reconstruction, directly resulting in watertight meshes.
The ability to accurately model cavities of the objects shows that our model uses texture information to improve over the visual hull (\figref{fig:mvs-visual-hull}).
Quantitatively, \tabref{tab:mvs} shows that our approach rivals the results from highly tuned MVS algorithms. 
We note that the DTU ground truth is itself sparse (\figref{fig:mvs-visual-hull-gt}) and methods are therefore rewarded for trading off completeness for accuracy, which explains the better quantitative performance of the baselines for higher trim parameters (\figref{fig:trim-parameter}).

\section{Conclusion and Future Work}

In this work, we have presented Differentiable Volumetric Rendering (DVR).
Observing that volumetric rendering is inherently differentiable for implicit representations allows us to formulate an analytic expression for the gradients of the depth with respect to the network parameters.
Our experiments show that DVR enables us to learn implicit 3D shape representations from multi-view imagery without 3D supervision, rivaling models that are learned with full 3D supervision.
Moreover, we found that our model can also be used for multi-view 3D reconstruction.
We believe that DVR is a useful technique that broadens the scope of applications of implicit shape and texture representations.

In the future, we plan to investigate how to circumvent the need for object masks and camera information, \eg, by predicting soft masks and how to estimate not only texture but also more complex material properties.

\section*{Acknowledgments}
This work was supported by an NVIDIA research gift.
The authors thank the International Max Planck Research School for Intelligent Systems (IMPRS-IS) for supporting Michael Niemeyer.

\FloatBarrier

{\small
\bibliographystyle{ieee_fullname}
\bibliography{bibliography_long,bibliography_cleaned,bibliography_custom}
}

\end{document}